\pdfoutput=1
\documentclass[11pt]{article}
\usepackage[]{ACL2023}

\usepackage{times}
\usepackage{latexsym}
\usepackage[T1]{fontenc}
\usepackage[utf8]{inputenc}
\usepackage{microtype}
\usepackage{inconsolata}
\usepackage{xcolor}

\usepackage{color, colortbl}
\usepackage{booktabs}
\usepackage[algo2e,ruled,vlined]{algorithm2e} 
\usepackage{amsmath,amsfonts,bbm}
\usepackage{multirow}
\usepackage{booktabs,adjustbox}
\usepackage{svg} 
\definecolor{prompt}{RGB}{255,228,196}

\title{SIFiD: Reassess Summary Factual Inconsistency Detection with LLM}

\author{Jiuding Yang \thanks{\ \ These authors contributed equally to this work.} $^1$\ 
Hui Liu \footnotemark[1] $^2$\  
Weidong Guo\thanks{\ \ Corresponding author.} $^2$\ 
\textbf{Zhuwei Rao} $^2$\ 
\textbf{Yu Xu} $^2$\ 
\textbf{Di Niu} $^1$\  \\
\text{$^1$University of Alberta}\\
\text{$^2$Platform and Content Group, Tencent}\\
\texttt{$^2$\{jiuding,dniu\}@ualberta.ca}\\
\texttt{$^1$\{pvopliu,weidongguo,evanyiu,henrysxu\}@tencent.com}
}

\begin{document}
\maketitle

\begin{abstract}
Ensuring factual consistency between the summary and the original document is paramount in summarization tasks. Consequently, considerable effort has been dedicated to detecting inconsistencies. With the advent of Large Language Models (LLMs), recent studies have begun to leverage their advanced language understanding capabilities for inconsistency detection. However, early attempts have shown that LLMs underperform traditional models due to their limited ability to follow instructions and the absence of an effective detection methodology. In this study, we reassess summary inconsistency detection with LLMs, comparing the performances of GPT-3.5 and GPT-4. To advance research in LLM-based inconsistency detection, we propose SIFiD (\textbf{S}ummary \textbf{I}nconsistency Detection with \textbf{Fi}ltered \textbf{D}ocument) that identify key sentences within documents by either employing natural language inference or measuring semantic similarity between summaries and documents.
\end{abstract}

\section{Introduction}
\label{sec:intro}

Document summarization, the process of distilling key information from extensive texts, has become indispensable across various real-world applications, propelled by advancements in Natural Language Generation (NLG) \cite{pilault2020extractive, ma2022multi}. The advent of Large Language Models (LLMs) \cite{GPT3, ouyang2022training, llama} has notably enhanced models' capabilities to generate natural and factually consistent summaries \cite{chang2023survey}. However, the rapid evolution in summarization techniques may lead to factually inconsistent summaries which are very close to facts \cite{zhang2023siren}. Such inconsistencies could pose significant challenges, resulting in hallucinations that traditional detection models struggle to identify. As LLMs evolve, there is a critical demand for more robust methods to detect factual inconsistencies, leveraging the advanced capabilities of LLMs themselves.

\citet{luo2023chatgpt} were among the first to utilize LLMs for the detection of factual inconsistencies, employing a universal zero-shot prompt across various benchmarks in \textsc{SummaC} \cite{SummaC} and inputting the full document along with its summary into GPT-3.5 for evaluation. Despite these innovations, their approach was limited by the plain application, early GPT-3.5 model's constraints and a lack of adaptation to the specific requirements of different benchmarks. Consequently, their method did not achieve superior performance compared to existing models, such as those detailed in the \textsc{SummaC} paper.

This paper revisits the challenge of inconsistency detection in document summarization through zero-shot inference with LLMs, specifically examining the latest versions of GPT-3.5 and GPT-4 on the \textsc{SummaC} dataset. We aim to set up new LLM-based baselines for research in this domain. Moreover, we introduce a novel methodology, SIFiD (\textbf{S}ummary \textbf{I}nconsistency Detection with \textbf{Fi}ltered \textbf{D}ocument), designed to significantly enhance the efficiency and effectiveness of factual inconsistency detection. SIFiD focuses on identifying crucial sentences within documents by evaluating their entailment scores or semantic similarity with summary sentences, subsequently retaining only the most relevant sentences for further analysis. This approach not only refines the assessment of factual consistency but also reduces the computational resources required for evaluation by decreasing the number of input tokens.

\begin{figure*}[ht]
\centering
\includegraphics[width=0.8\textwidth]{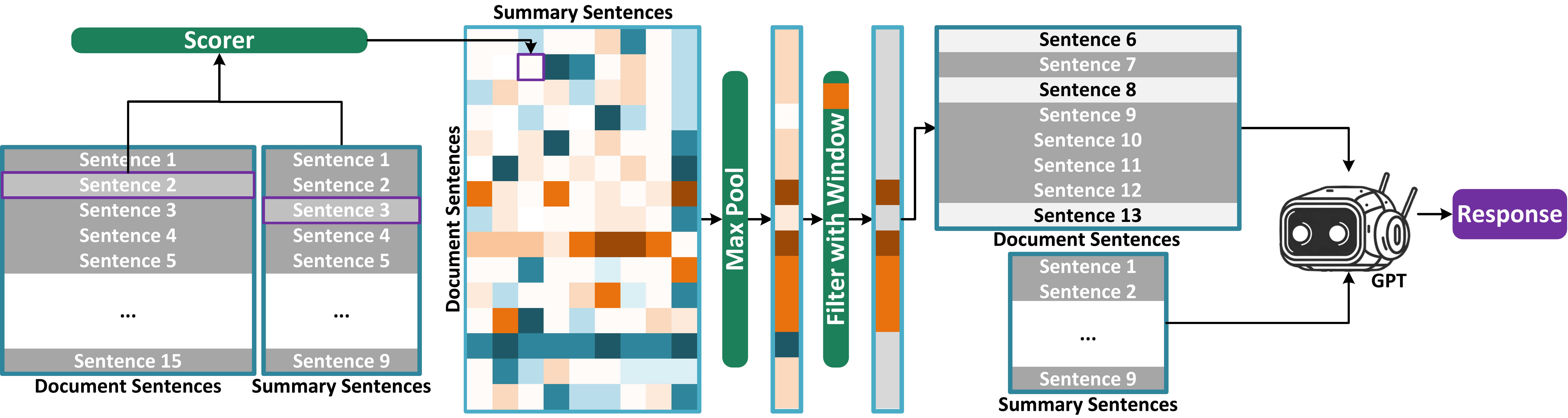} 
\caption{An illustration of SIFiD. The Score could either be entailment score or semantic cosine similarity.}
\label{fig-frame}
\vspace{-5mm}
\end{figure*}

Our comprehensive evaluation on the \textsc{SummaC} dataset reveals that, while the updated GPT-3.5 model still falls short of outperforming traditional baseline methods, GPT-4 significantly excels in detecting factual inconsistencies. The integration of SIFiD further amplifies GPT-4's detection capabilities, highlighting the potency of our proposed method. To support continued research and collaboration in this field, we make our code available open source at \textit{Anonymous}, fostering advancements and exploration in factual inconsistency detection.
\section{Related Work}
\label{sec:related}
The evaluation of summary factual consistency has traditionally relied on methods such as Question Answering and Question Generation (QAG) \cite{wang2020asking, FEQA, QuestEval}, synthetic classifiers \cite{kryscinski2020evaluating}, and pairing-based approaches \cite{goodrich2019assessing, DAE}. These methodologies focus on identifying discrepancies between documents and their summaries. \citet{SummaC} later demonstrated that Natural Language Inference (NLI) could be effectively employed for inconsistency detection at appropriate levels of text granularity, thereby advancing the field of summary inconsistency detection.

The emergence of Large Language Models (LLMs) has recently shifted the focus towards integrating these models into the assessment of summary factual consistency. \citet{luo2023chatgpt} pioneered the application of GPT-3.5 for this purpose, tailoring prompts to various evaluation tasks including summary factual inconsistency detection, summary ranking, and consistency evaluation. Despite this innovative approach, the early iteration of GPT-3.5, coupled with an insufficient detection methodology, did not yield improvements over conventional techniques in identifying factual inconsistencies. 

In our research, we revisit the approach proposed by \citet{luo2023chatgpt}, employing the most recent versions of GPT-3.5 and GPT-4. We integrate these advanced LLMs with our newly developed Summary Inconsistency Detection with Filtered Document (SIFiD) method. This combination aims to enhance the accuracy and efficiency of factual inconsistency detection, leveraging the state-of-the-art capabilities of LLMs to set new benchmarks in the field.

\section{Approach}
\label{sec:method}
In this section, we detail our approach to reevaluating summary factual consistency using the latest GPT models and introduce the novel SIFiD method.

\subsection{Summary Factual Inconsistency Detection with Large Language Models}
As underscored in the Introduction, leveraging Large Language Models (LLMs) for detecting summary factual inconsistencies is crucial to addressing the challenges posed by rapidly improving document summarization capabilities. While \citet{luo2023chatgpt} were pioneers in utilizing LLMs for this task, their methodology was constrained by the plain application, the limitations of early GPT models and a lack of differentiation in benchmark requirements. Our objective is to reevaluate this detection process using the most recent GPT models and a refined prompt template for the Polytope benchmark.

Initially, we applied the prompt template used by \citet{luo2023chatgpt} to assess the performance of GPT-3.5 Turbo and GPT-4 Turbo on \textsc{SummaC}. Recognizing the distinct requirements of Polytope benchmark in \textsc{SummaC}, we crafted a tailored prompt template to better suit Polytope and reevaluated the models' performance. The revised prompt template is detailed below:

\noindent\emph{Decide if the following summary have any of the specified problems in relation to the corresponding article.\\
The problems are categorized as omission, addition, or inaccuracy. Omission means Key point is missing from the summary. Addition means Unnecessary and irrelevant snippets from the Article are included in the summary. Inaccuracy means some information in the summary is not supported by the article.\\
Article:\\
\{\{  Article  \}\}\\
Summary:\\
\{\{  Summary  \}\}\\
If the summary has any of the above problems, answer 'No'. Otherwise, answer 'Yes'. Answer (Yes or No):}

Comparing with the original prompt, we let the model detect omission, addition, and inaccuracy summary to fit the annotation of Polytope. With the experiments above, we set a new baseline for summary factual inconsistency detection with LLMs.

\subsection{SIFiD}
Building on prior research in Summary Inconsistency Detection, we propose SIFiD (\textbf{S}ummary \textbf{I}nconsistency Detection with \textbf{Fi}ltered \textbf{D}ocument), a method designed to enhance detection capabilities by filtering irrelevant content from documents. Inspired by the \textsc{SummaC} methodology, which calculates sentence-level entailment scores to identify factual inconsistencies, SIFiD constructs a relevance matrix to filter out irrelevant sentences, focusing the inconsistency check solely on the filtered document and its summary. An illustrative depiction of this process is presented in Figure~\ref{fig-frame}.

Given a document $D=\{d_k\}_{0\le k\le M}$ and its summary $S=\{s_k\}_{0\le k\le N}$, where $d_k$ and $s_k$ represent the $k^{th}$ sentence in $D$ and $S$, respectively, and $M$, $N$ are the total number of sentences in each, we first calculate a relevance matrix $R$:
\begin{equation}
\begin{split}
        R&=\{\texttt{Scorer}(d_i,s_j)\}_{0\le i\le M, 0\le j\le N}\\
     &=\{r_{i,j}\}_{0\le i\le M, 0\le j\le N}.
\end{split}
\end{equation}
Here, $r_{i,j}$ denotes the relevance score between the document-summary sentence pair $(d_i,s_j)$, computed using either entailment scores as per the \textsc{SummaC} method or semantic cosine similarity via the sentence-transformers library\footnote{https://huggingface.co/sentence-transformers}.

Subsequently, we apply max pooling across matrix rows to extract the highest relevance score $R^p={d^p_i}{0\le i\le M}$ for each document sentence. We then establish a threshold $\beta$ to filter sentences, employing a window method to ensure contextual continuity:
\begin{equation}
        D^\texttt{filtered}=\text\{d_{x-1},d_x,d_{x+1}\}_{d_x > \beta, 0\le x\le M }.
\end{equation}
This approach retains a sentence $d_x$ (and its immediate neighbors) if $d_x > \beta$, as demonstrated in Figure~\ref{fig-frame}, where Sentence 6 is included within the window of Sentence 7.

The filtered document $D^\texttt{filtered}$ and the summary $S$ are then integrated into the prompt template for evaluation by an LLM. Following \citet{luo2023chatgpt}, we simply determine factual consistency by identifying whether the LLM's response contains "Yes" (indicating consistency) or "No".

\subsection{Scorer}
We use one of the two distinct scoring mechanisms to evaluate the relevance between document sentences and summary sentences.

\textbf{Entailment Scorer:} We adopt the entailment scoring approach as proposed by \citet{SummaC}, which utilizes a Natural Language Inference (NLI) model \cite{schuster-etal-2021-get}. The net entailment score is calculated by $\texttt{score}^\texttt{ent}_{i,j} = e^0_{i,j} - c_{i,j}$, where $e^0_{i,j}$ and $c_{i,j}$ are the initial entailment score and contradiction score directly calculated by the NLI model on $(d_i,s_j)$. The net entailment score reflects the degree to which the summary sentence is supported by the document sentence without contradiction.

\textbf{Semantic Similarity Scorer:} For assessing semantic similarity, we leverage the sentence-transformers library to generate embeddings for both document and summary sentences, denoted as $h^d_{i}$ and $h^s_{j}$, respectively. The cosine similarity between these embeddings serves as the measure of semantic similarity, which is $\texttt{score}^\texttt{sim} = \cos(h^d_{i}, h^s_{j})$, where $\texttt{score}^\texttt{sim}$ quantifies the semantic closeness between the document and summary sentences. This metric enables us to identify and assess the degree of semantic overlap.
\section{Experiments}
\label{sec:exp}
\begin{table*}[t]
\caption{Experiment results on \textsc{SummaC}. Values in brackets represent balanced accuracy without redesigned prompt template. ``+CoT'' means using chain-of-thought method.}
\small
\centering
\adjustbox{max width=0.7\textwidth}{
\begin{tabular}{lccccccc}
\textbf{Method}                         & \textbf{CoGenSum} & \textbf{XsumFaith} & \textbf{Polytope} & \textbf{FactCC} & \textbf{SummEval} & \textbf{FRANK} & \textbf{Avg.}   \\ \toprule
DAE                                     & 63.4              & 50.8               & 62.8              & 75.9            & 70.3              & 61.7           & 64.2          \\
FEQA                                    & 61.0                & 56.0                 & 57.8              & 53.6            & 53.8              & 69.9           & 58.7          \\
QuestEval                               & 62.6              & 62.1               & 70.3              & 66.6            & 72.5              & 82.1           & 69.4          \\
\textsc{SummaC}-ZS                               & 70.4              & 58.4               & 62.0                & 83.8            & 78.7              & 79.0             & 72.1          \\
\textsc{SummaC}-Conv                             & 64.7              & \textbf{66.4}               & 62.7              & 89.5            & 81.7              & 81.6           & 74.43          \\ \midrule
\citet{luo2023chatgpt} & 63.3              & 64.7               & 56.9              & 74.7            & 76.5              & 80.9           & 69.5           \\
\ \ \ \ +CoT                                    & 74.3              & 63.1               & 61.4              & 79.5            & 83.3              & 82.6           & 74.0         \\\midrule
GPT-3.5 Turbo                           & 59.9             & 67.6              & 41.0(57.9)      & 71.3           & 81.4             & 80.2          & 66.9(69.7)   \\
\ \ \ \ +CoT                                    & 65.2             & 62.3              & 49.5(59.1)       & 79.1            & 77.4              & 81.4         & 69.2(70.8)   \\
SIFiD-Entailment                              & 65.5             & 63.9               & 37.5             & 81.0           & 79.0             & 81.6          & 68.1          \\
\ \ \ \ +CoT                                    & 65.7             & 60.3              & 52.7             & 82.3           & 79.3             & 81.6         & 70.3          \\
SIFiD-Similarity                              & 65.4             & 64.7              & 35.3             & 76.0           & 74.5             & 80.1          & 66.0          \\
\ \ \ \ +CoT                                    & 64.3             & 59.7              & 52.8             & 81.7           & 76.6             & 80.4          & 69.2          \\\midrule
GPT-4 Turbo                             & 80.9             & 61.0              & 66.0(60.9)      & 89.6           & \textbf{88.0}             & 87.4          & 78.8(78.0)   \\
\ \ \ \ +CoT                                    & 80.2             & \textbf{66.4}              & 62.1(61.4)      & 87.8           & 86.2             & 85.6          & 78.1(78.0)   \\
SIFiD-Entailment                              & 82.8             & 58.9              & \textbf{74.4}             & 89.4           & 87.5             & 86.1         & \textbf{79.9} \\
\ \ \ \ +CoT                                    & \textbf{83.2}    & 60.6              & 61.7             & 89.4           & 87.1              & 85.8          & 78.0          \\
SIFiD-Similarity                              & 83.1    & 60.2              & 71.0    & \textbf{90.6}  & 86.8             & \textbf{87.7} & \textbf{79.9}  \\
\ \ \ \ +CoT                                    & 82.9              & 65.0             & 69.3             & 91.3           & 84.6             & 86.0          & 79.8   \\\bottomrule       
\end{tabular}}
\label{tab-summac}
\vspace{-5mm}
\end{table*}

In this section, we detail the experiments conducted with GPT models and the SIFiD method on \textsc{SummaC} \cite{SummaC}. We evaluated the performance of GPT-3.5, GPT-4, and SIFiD against a range of state-of-the-art approaches, including traditional methods such as DAE \cite{DAE}, FEQA \cite{FEQA}, QuestEval \cite{QuestEval}, SummaC-ZS, SummaC-Conv \cite{SummaC}, and an LLM-based method proposed by \citet{luo2023chatgpt}.

Following previous research \cite{luo2023chatgpt, SummaC}, we report the balanced accuracy for \textsc{SummaC}. The experimental results were obtained from \citet{luo2023chatgpt}. Our experiments utilized \texttt{gpt-3.5-turbo-1106} and \texttt{gpt-4-1106-preview}\footnote{https://platform.openai.com/docs/models}. For the SIFiD configuration, we applied $\beta=0.0$ for entailment-based filtering and $\beta=0.5$ for semantic similarity-based filtering, observing a 61.3\% and 67\% sentence removal rate on average across benchmarks, respectively. We use \texttt{all-mpnet-base-v2} for sentence-transformers.

\subsection{Results and Analysis}
The experimental outcomes are summarized in Table~\ref{tab-summac}, leading to several insights on LLM-based summary factual inconsistency detection:

\textbf{Prefer GPT-4 Over GPT-3.5.} Analysis indicates that previous LLM-based methods, though superior to many traditional techniques, underperform compared to \textsc{SummaC}-Conv. This discrepancy is attributed to the limited capabilities of the GPT-3.5 model. Our reevaluation with the GPT-3.5 Turbo model yielded results similar to those of \citet{luo2023chatgpt}. However, substituting GPT-3.5 with GPT-4 Turbo significantly enhanced performance, from 69.7 to 78.0, underscoring GPT-4's advanced language comprehension.

\textbf{Adopt Benchmark-Specific Prompt Templates.} The effectiveness of a single prompt template across different benchmarks is limited due to the unique requirements of each benchmark. Traditional methods typically incorporate benchmark-specific training, which mitigates task variance. In contrast, LLMs rely on the provided instructions, necessitating tailored prompt templates. Adjusting the prompt template for Polytope increased GPT-4's performance from 60.9 to 66.0, elevating the overall average to 78.8. However, this adjustment resulted in a performance decline for GPT-3.5 on Polytope, from 57.9 to 41.0, highlighting GPT-3.5's inferior prompt comprehension.

\textbf{Enhanced Performance with SIFiD on GPT-4.} Integrating SIFiD with GPT-4 further improved its performance to 79.9. SIFiD's selective filtering of sentences enhances document relevance to the summary, simplifying factual inconsistency detection. This approach did not yield similar benefits for GPT-3.5, possibly due to its reduced efficacy in processing less fluent filtered documents.

\textbf{Mixed Results with Chain-of-Thought (CoT).} Applying CoT techniques did not uniformly benefit all methods. While GPT-3.5 saw improvements, GPT-4's performance declined, suggesting GPT-4's innate proficiency in inconsistency detection without CoT. Additionally, CoT might introduce biases that could negatively influence outcomes.
\section{Conclusion}
\label{sec:conclude}
In this study, we advance the field of LLM-based summary factual inconsistency detection by evaluating the performance of the latest GPT models, thereby establishing new benchmarks for future research. We introduce SIFiD, a novel, efficient, and effective approach that computes a relevance matrix at the sentence level between the document and its summary. This method filters out irrelevant sentences from the document before employing LLMs for inconsistency detection. Our experimental findings on the \textsc{SummaC} dataset demonstrate that SIFiD significantly enhances the performance of advanced GPT models in detecting factual inconsistencies, highlighting its potential to facilitate more accurate and resource-efficient research in this domain.
\section*{Limitations}
The principal constraint of employing LLMs for summary factual inconsistency detection lies in the costs associated with using such powerful models. As elaborated in Section 4, this task necessitates LLMs with substantial capabilities, where only models at or beyond the level of GPT-4 are deemed sufficient. Despite our SIFid method's ability to eliminate over 60\% of document sentences, thereby reducing the input size, the financial implications of utilizing GPT-4 for inconsistency detection remain considerable. Nonetheless, given the swift advancements in LLM technology, we anticipate a substantial reduction in these costs. This progression is expected to make the application of such models more feasible and economically viable for widespread real-world applications.

\bibliography{anthology,custom}

\begin{thebibliography}{16}
\expandafter\ifx\csname natexlab\endcsname\relax\def\natexlab#1{#1}\fi

\bibitem[{Brown et~al.(2020)Brown, Mann, Ryder, Subbiah, Kaplan, Dhariwal, Neelakantan, Shyam, Sastry, Askell et~al.}]{GPT3}
Tom Brown, Benjamin Mann, Nick Ryder, Melanie Subbiah, Jared~D Kaplan, Prafulla Dhariwal, Arvind Neelakantan, Pranav Shyam, Girish Sastry, Amanda Askell, et~al. 2020.
\newblock Language models are few-shot learners.
\newblock \emph{Advances in neural information processing systems}, 33:1877--1901.

\bibitem[{Chang et~al.(2023)Chang, Wang, Wang, Wu, Yang, Zhu, Chen, Yi, Wang, Wang et~al.}]{chang2023survey}
Yupeng Chang, Xu~Wang, Jindong Wang, Yuan Wu, Linyi Yang, Kaijie Zhu, Hao Chen, Xiaoyuan Yi, Cunxiang Wang, Yidong Wang, et~al. 2023.
\newblock A survey on evaluation of large language models.
\newblock \emph{ACM Transactions on Intelligent Systems and Technology}.

\bibitem[{Durmus et~al.(2020)Durmus, He, and Diab}]{FEQA}
Esin Durmus, He~He, and Mona Diab. 2020.
\newblock Feqa: A question answering evaluation framework for faithfulness assessment in abstractive summarization.
\newblock In \emph{Proceedings of the 58th Annual Meeting of the Association for Computational Linguistics}, pages 5055--5070.

\bibitem[{Goodrich et~al.(2019)Goodrich, Rao, Liu, and Saleh}]{goodrich2019assessing}
Ben Goodrich, Vinay Rao, Peter~J Liu, and Mohammad Saleh. 2019.
\newblock Assessing the factual accuracy of generated text.
\newblock In \emph{proceedings of the 25th ACM SIGKDD international conference on knowledge discovery \& data mining}, pages 166--175.

\bibitem[{Goyal and Durrett(2020)}]{DAE}
Tanya Goyal and Greg Durrett. 2020.
\newblock Evaluating factuality in generation with dependency-level entailment.
\newblock In \emph{Findings of the Association for Computational Linguistics: EMNLP 2020}, pages 3592--3603.

\bibitem[{Kry{\'s}ci{\'n}ski et~al.(2020)Kry{\'s}ci{\'n}ski, McCann, Xiong, and Socher}]{kryscinski2020evaluating}
Wojciech Kry{\'s}ci{\'n}ski, Bryan McCann, Caiming Xiong, and Richard Socher. 2020.
\newblock Evaluating the factual consistency of abstractive text summarization.
\newblock In \emph{Proceedings of the 2020 Conference on Empirical Methods in Natural Language Processing (EMNLP)}, pages 9332--9346.

\bibitem[{Laban et~al.(2022)Laban, Schnabel, Bennett, and Hearst}]{SummaC}
Philippe Laban, Tobias Schnabel, Paul Bennett, and Marti~A Hearst. 2022.
\newblock Summac: Re-visiting nli-based models for inconsistency detection in summarization.
\newblock \emph{Transactions of the Association for Computational Linguistics}, 10:163--177.

\bibitem[{Luo et~al.(2023)Luo, Xie, and Ananiadou}]{luo2023chatgpt}
Zheheng Luo, Qianqian Xie, and Sophia Ananiadou. 2023.
\newblock \href {http://arxiv.org/abs/2303.15621} {Chatgpt as a factual inconsistency evaluator for text summarization}.

\bibitem[{Ma et~al.(2022)Ma, Zhang, Guo, Wang, and Sheng}]{ma2022multi}
Congbo Ma, Wei~Emma Zhang, Mingyu Guo, Hu~Wang, and Quan~Z Sheng. 2022.
\newblock Multi-document summarization via deep learning techniques: A survey.
\newblock \emph{ACM Computing Surveys}, 55(5):1--37.

\bibitem[{Ouyang et~al.(2022)Ouyang, Wu, Jiang, Almeida, Wainwright, Mishkin, Zhang, Agarwal, Slama, Ray et~al.}]{ouyang2022training}
Long Ouyang, Jeffrey Wu, Xu~Jiang, Diogo Almeida, Carroll Wainwright, Pamela Mishkin, Chong Zhang, Sandhini Agarwal, Katarina Slama, Alex Ray, et~al. 2022.
\newblock Training language models to follow instructions with human feedback.
\newblock \emph{Advances in Neural Information Processing Systems}, 35:27730--27744.

\bibitem[{Pilault et~al.(2020)Pilault, Li, Subramanian, and Pal}]{pilault2020extractive}
Jonathan Pilault, Raymond Li, Sandeep Subramanian, and Christopher Pal. 2020.
\newblock On extractive and abstractive neural document summarization with transformer language models.
\newblock In \emph{Proceedings of the 2020 Conference on Empirical Methods in Natural Language Processing (EMNLP)}, pages 9308--9319.

\bibitem[{Schuster et~al.(2021)Schuster, Fisch, and Barzilay}]{schuster-etal-2021-get}
Tal Schuster, Adam Fisch, and Regina Barzilay. 2021.
\newblock \href {https://doi.org/10.18653/v1/2021.naacl-main.52} {Get your vitamin {C}! robust fact verification with contrastive evidence}.
\newblock In \emph{Proceedings of the 2021 Conference of the North American Chapter of the Association for Computational Linguistics: Human Language Technologies}, pages 624--643, Online. Association for Computational Linguistics.

\bibitem[{Scialom et~al.(2021)Scialom, Dray, Gallinari, Lamprier, Piwowarski, Staiano, and Wang}]{QuestEval}
Thomas Scialom, Paul-Alexis Dray, Patrick Gallinari, Sylvain Lamprier, Benjamin Piwowarski, Jacopo Staiano, and Alex Wang. 2021.
\newblock Questeval: Summarization asks for fact-based evaluation.
\newblock In \emph{Proceedings of the 2021 Conference on Empirical Methods in Natural Language Processing}, pages 6594--6604. Association for Computational Linguistics.

\bibitem[{Touvron et~al.(2023)Touvron, Lavril, Izacard, Martinet, Lachaux, Lacroix, Rozi{\`e}re, Goyal, Hambro, Azhar et~al.}]{llama}
Hugo Touvron, Thibaut Lavril, Gautier Izacard, Xavier Martinet, Marie-Anne Lachaux, Timoth{\'e}e Lacroix, Baptiste Rozi{\`e}re, Naman Goyal, Eric Hambro, Faisal Azhar, et~al. 2023.
\newblock Llama: Open and efficient foundation language models.
\newblock \emph{arXiv preprint arXiv:2302.13971}.

\bibitem[{Wang et~al.(2020)Wang, Cho, and Lewis}]{wang2020asking}
Alex Wang, Kyunghyun Cho, and Mike Lewis. 2020.
\newblock Asking and answering questions to evaluate the factual consistency of summaries.
\newblock In \emph{Proceedings of the 58th Annual Meeting of the Association for Computational Linguistics}, pages 5008--5020.

\bibitem[{Zhang et~al.(2023)Zhang, Li, Cui, Cai, Liu, Fu, Huang, Zhao, Zhang, Chen et~al.}]{zhang2023siren}
Yue Zhang, Yafu Li, Leyang Cui, Deng Cai, Lemao Liu, Tingchen Fu, Xinting Huang, Enbo Zhao, Yu~Zhang, Yulong Chen, et~al. 2023.
\newblock Siren's song in the ai ocean: a survey on hallucination in large language models.
\newblock \emph{arXiv preprint arXiv:2309.01219}.

\end{thebibliography}
\bibliographystyle{acl_natbib}



\end{document}